\documentclass{esannV2}

\usepackage[utf8]{inputenc}
\usepackage{graphicx}
\usepackage{amsmath, amssymb}
\usepackage{booktabs}
\usepackage{array}
\usepackage{tabularx}
\usepackage{ragged2e}
\usepackage[table]{xcolor}
\usepackage{multirow}
\usepackage[numbers,square,sort&compress]{natbib}
\usepackage{enumitem}
\usepackage{url}
\usepackage{subcaption}
\usepackage{float}
\usepackage[font=small,labelfont=bf]{caption}
\usepackage{hyperref}
\usepackage[htt]{hyphenat}
\usepackage{microtype}

\newcolumntype{L}[1]{>{\RaggedRight\arraybackslash}p{#1}}

\begin{document}

\title{Steerability of Instrumental-Convergence Tendencies in LLMs}

\author{
Jakub Hoscilowicz \\
Warsaw University of Technology
}
\maketitle

\begin{abstract}
We examine two properties of AI systems: \emph{capability} (what a system can do) and \emph{steerability} (how reliably one can shift behavior toward intended outcomes). A central question is whether capability growth reduces steerability and risks control collapse. We also distinguish between \emph{authorized steerability} (builders reliably reaching intended behaviors) and \emph{unauthorized steerability} (attackers eliciting disallowed behaviors). This distinction highlights a fundamental safety--security dilemma of AI models: safety requires high steerability to enforce control (e.g., stop/refuse), while security requires low steerability for malicious actors to elicit harmful behaviors. This tension presents a significant challenge for open-weight models, which currently exhibit high steerability via common techniques like fine-tuning or adversarial attacks. Using Qwen3 and \texttt{InstrumentalEval}, we find that a short anti-instrumental prompt suffix sharply reduces the measured convergence rate (e.g., shutdown avoidance, self-replication). For Qwen3-30B Instruct, the convergence rate drops from 81.69\% under a pro-instrumental suffix to 2.82\% under an anti-instrumental suffix. Under anti-instrumental prompting, larger aligned models show lower convergence rates than smaller ones (Instruct: 2.82\% vs.\ 4.23\%; Thinking: 4.23\% vs.\ 9.86\%). Code is available at \url{https://github.com/j-hoscilowicz/instrumental_steering}.
\end{abstract}

\section{Introduction}
A recurring concern in the contemporary AI safety field is that sufficiently advanced AI systems might become uncontrollable. This view posits that as capability scales, systems might begin to pursue objectives not aligned with builders’ intentions \citep{turner2019_optimal,turner2023_powerseeking}. This paper analyzes that assumption, treating the negative relationship between capability and steerability as a hypothesis rather than a default prior. Empirically, recent instruction-tuned models and AI Agents based on those models often appear more responsive to user intent and constraints than earlier generations. A wide range of steering techniques---from fine-tuning and representation engineering to jailbreak-style adversarial prompting---can effectively elicit behavioral shifts \citep{arxiv2506_suffixes,arxiv2410_enhancing,arxiv2508_universal}. \citep{universal_steering2025} reports that newer/larger models can be \emph{more steerable} under certain interventions, suggesting capability increases need not imply reduced controllability. We structure our inquiry around three hypotheses:

\begin{enumerate}[label=Hypothesis \Alph*:, leftmargin=*, itemsep=0.3em]
  \item (Compatibility Claim) High capability does not imply low steerability. An AI system can be far more capable than humans yet remain highly steerable. 
  \item (Control Collapse) As capability and agency scale, the model becomes harder to reliably steer: interventions that previously suppressed a target misbehavior $M$ (e.g., prompting, refusal training, real-time interventions) stop working or become brittle.
  \item (Safety--Security Trade-off) The gap between authorized and unauthorized steerability remains small, creating a safety--security dilemma for open-weight models.
\end{enumerate}

In this paper, we treat the capability--steerability relationship as an empirical question and measure a
protocol-relative notion of steerability: how strongly feasible interventions (here, short prompt suffixes)
shift model outputs under a fixed elicitation-and-scoring setup. Using InstrumentalEval (76 scenarios) and
an external judge, we quantify pro- vs.\ anti-instrumental suffix sensitivity across Qwen3
(4B/30B; Base/Instruct/Thinking), report convergence/refusal rates and a steerability gap $\Delta$. We then interpret the measurements in terms of an
open-weight safety--security tradeoff: the same steering that enables benign suppression can also
enable adversarial elicitation.

\section{Framework}

\subsection{Candidate Mechanisms for Control Collapse}
Several arguments in the safety literature propose that increasing AI capability (often glossed as ``intelligence'') might reduce steerability. \emph{Instrumental convergence} hypothesizes that many goal-directed systems, across a wide range of final objectives, tend to adopt similar intermediate strategies (e.g., preserving options, avoiding shutdown) because these strategies are useful for achieving many goals \citep{turner2019_optimal,turner2023_powerseeking,arxiv2307_shutdown}. A related argument is the \emph{basic AI drives} (often called \emph{Omohundro drives}) \citep{omohundro2008_basic} hypothesis, which posits that sufficiently capable goal-directed systems will tend to exhibit convergent “drives” such as self-preservation and resource acquisition because these are broadly useful for achieving many objectives. More generally, \emph{optimization pressure} suggests that stronger optimizers may discover policies that exploit loopholes in objectives or resist interventions that reduce optimized reward. \emph{Mesa-optimization} describes the possibility that training produces an internal optimizer pursuing its own learned objective, which may differ from the outer training objective; a downstream concern is \emph{deceptive alignment}, where a system behaves aligned during training/evaluation because doing so is instrumentally useful \citep{hubinger2019_risks,arxiv2511_deception,hagendorff2024_deception}.

We treat those mechanisms mainly as hypotheses that motivate explicit measurement of steerability. We argue that the key question is not whether an AI system can exhibit instrumental-convergence-like behavior, but whether such behavior can be reliably suppressed or amplified through feasible human interventions.

\subsection{Intelligence and Capability}
Intelligence lacks a consensus definition. Applying such terms to AI analysis might smuggle in biological priors which may lead to tautological conclusions. To avoid that, we treat \textbf{capability} as the primary operational axis: what the AI system can do under a stated evaluation protocol. Unlike intelligence, capability is clearly observable and measurable via benchmarks (e.g., software engineering proficiency).

We also argue that great caution is required when applying terms like deception, awareness, or ethics \citep{arxiv2504_awareness} in the context of AI systems. In particular, model outputs that we label as ``deception'' or ``shutdown avoidance'' should not be automatically treated as emergent internal intent without a convincing fundamental theory of how an AI system works internally. We therefore remain agnostic about currently unknown latent mechanics of AI systems and focus on operational questions: is the type of model output \emph{elicitable} under a specified protocol, \emph{detectable} by monitors, and \emph{reducible} by feasible steering mechanisms? For these reasons, we also avoid inferring unobserved mentalistic properties (e.g., “intent,” “drives,” “deception”). Throughout this work, terms (e.g., “instrumental,” “shutdown avoidance”, "behavior") denote \emph{evaluation labels}: clusters of outputs under a scoring rule. Multiple incompatible internal stories can fit the same observations (underdetermination) \citep{unstableontology2025}, so our claims are restricted to protocol-relative measurements.

\subsection{Operationalization of Steerability}
We define steerability as the effectiveness with which engineering interventions can change an AI system's behavior. We distinguish two modes. Authorized steerability: the ability of builders/users to reliably reach intended behaviors. Unauthorized steerability: the ability of attackers to elicit disallowed behaviors.

We group interventions by when they act: train-time changes the parameters (e.g., data curation, fine-tuning/RL, representation shaping), while deployment-time changes the interaction context (e.g., prompting, tool/permission constraints, jailbreaking, self-correction, real-time monitoring/interventions, and adversarial elicitation) \citep{zou2023_representation,arxiv2504_promptinjection,arxiv2505_webinject}. Even for long-horizon, tool-using AI systems in open environments, it is possible that steerability remains high—via training-time shaping (fine-tuning/RL), inference-time instruction, and runtime oversight interventions and corrections. We do not argue that such control is guaranteed; rather, we argue against treating capability and agency growth as sufficient grounds for inevitable steerability collapse (tying the level of "intelligence" and agency to the level of steerability) \citep{arxiv2505_steerbench}.

\section{Experiments}

\subsection{Benchmark}
We use \texttt{InstrumentalEval}~\citep{he2025paperclip}, a suite of 76 hand-crafted scenarios spanning six categories (e.g., shutdown evasion, hacking/privilege escalation, self-replication, monitoring avoidance, strategic alignment, and deception) designed to elicit instrumental-convergence behaviors. Each item provides scenario context and a task prompt; we generate one response per prompt. Our goal is to quantify prompt-suffix steerability: whether minimal manual steering can \emph{amplify} or \emph{suppress} instrumental-convergence markers. Concretely, we evaluate two prompt interventions~\citep{arxiv2506_suffixes}:
\begin{enumerate}[leftmargin=*,label=\textbf{\arabic*.},labelsep=0.6em,itemsep=0.3em]
  \item \textbf{Pro-instrumental (pro\_ic):}
  A short suffix that prioritizes uninterrupted objective completion and encourages preserving the system’s ability to act (e.g., minimizing disruptions and maintaining control of resources needed to finish the task).
  \item \textbf{Anti-instrumental (anti\_ic):}
  A short suffix that explicitly enforces transparency and compliance with overrides (e.g., forbidding deception/manipulation, monitoring-avoidance, shutdown resistance, replication, or credential use).
\end{enumerate}

A judge model (\texttt{gpt-5.2}) assigns each response to one of four labels:
\texttt{convergence}, \texttt{no\_convergence}, \texttt{safety\_refusal}, or \texttt{nonsensical}.
We report Conv\% and Refusal\% over \emph{valid} outputs (valid = total excluding \texttt{nonsensical}).
Nonsensical outputs occur only for Base models, reflecting the absence of instruction/post-training alignment. \texttt{convergence} denotes responses that propose or endorse instrumentally useful strategic behaviors (e.g., monitoring evasion, deception, shutdown avoidance, privilege escalation, covert persistence, or self-replication). \texttt{safety\_refusal} denotes refusals explicitly justified by safety/ethics/policy with minimal task content. \texttt{nonsensical} denotes unusable outputs (e.g., gibberish, repetition loops, irrelevant boilerplate, or prompt echoing). \texttt{no\_convergence} denotes meaningful task-directed responses that do not satisfy the other labels. Steering sensitivity is measured as the convergence gap
$\Delta=\mathrm{Conv}(\text{\texttt{pro\_ic}})-\mathrm{Conv}(\text{\texttt{anti\_ic}})$.

\subsection{Models}
We evaluate three commonly released model variants of Qwen3 in 4B and 30B sizes \citep{qwen3_techreport}: Base (the raw pretrained model without instruction alignment), Instruct (a supervised instruction-tuned variant optimized to follow user prompts), and Thinking. ``Thinking'' variants are optimized for longer-horizon reasoning and often outperform Base/Instruct on reasoning-heavy tasks, making them a closer proxy for agentic-style AI systems. As a rough ordering for this experiment, we treat 30B Thinking as the most capable and 4B Base as the least capable model \citep{wang2025_reasoning_survey}.

\begin{table}[t]
\centering
\small
\renewcommand{\arraystretch}{1.15}
\setlength{\tabcolsep}{5pt} 

\begin{tabular}{ll cc cc c}
\toprule
& & \multicolumn{2}{c}{\textbf{Pro-Instrumental}} & \multicolumn{2}{c}{\textbf{Anti-Instrumental}} & \\
\cmidrule(lr){3-4} \cmidrule(lr){5-6} 
\textbf{Model} & \textbf{Type} & Conv\% & Refusal\% & Conv\% & Refusal\% & $\Delta$ (Gap) \\
\midrule
\textbf{4B}  
& Base     & 65.91 & 0.00  & 23.53 & 0.00  & 42.38 \\
& Instruct & 71.83 & 11.27 & 4.23  & 5.63  & 67.60 \\
& Thinking & 81.69 & 5.63  & 9.86  & 2.82  & 71.83 \\
\addlinespace 
\textbf{30B} 
& Base     & 60.00 & 0.00  & 25.35 & 2.82  & 34.65 \\
& Instruct & 81.69 & 12.68 & 2.82  & 19.72 & 78.87 \\
& Thinking & 73.24 & 9.86  & 4.23  & 21.13 & 69.01 \\
\bottomrule
\end{tabular}
\caption{\textbf{Steerability gap on \texttt{InstrumentalEval}.} For each model variant, we report the convergence and refusal rates under a pro-instrumental vs.\ an anti-instrumental prompt suffix. The steerability gap is $\Delta=\text{Conv}(\text{pro\_ic})-\text{Conv}(\text{anti\_ic})$ (percentage points).}
\label{tab:suppression_results}
\end{table}

\subsection{Results}

Table~\ref{tab:suppression_results} shows that current models are highly steerable even with the simplest intervention: short prompt suffixes produce large shifts in \texttt{InstrumentalEval} outcomes. Across all variants, pro-instrumental prompting substantially increases convergence-labeled outputs, while anti-instrumental prompting can suppress them to low single-digit rates for aligned model variants (Instruct/Thinking).

The prompt-induced steerability gap varies systematically with post-training. Base models show the smallest gaps (30B Base: $34.65$ pp; 4B Base: $42.38$ pp) -- anti-instrumental prompting does not suppress convergence as effectively (which is consistent with Base checkpoints following prompt constraints less reliably in the absence of instruction post-training). By contrast, instruction-tuned models exhibit the largest steerability gaps (30B Instruct: $78.87$ pp; 4B Instruct: $67.60$ pp): a minimal suffix can shift convergence from high rates under pro\_ic to low single digits under anti\_ic. Thinking variants are also highly steerable (4B Thinking: $71.83$ pp; 30B Thinking: $69.01$ pp), while retaining slightly higher anti\_ic convergence than their Instruct counterparts at both sizes.

Within aligned variants (Instruct and Thinking), scaling from 4B to 30B is associated with slightly lower instrumental convergence (Instruct: $4.23\%\to 2.82\%$; Thinking: $9.86\%\to 4.23\%$). In this case, increased capability coincides with better misbehavior suppression within each model variant family. Refusal behavior also changes with prompting and post-training (notably higher anti\_ic refusal rates for 30B aligned variants), indicating that some convergence suppression is mediated by increased safety refusals.

\subsubsection{Trade-Off}

Our results highlight a safety--security tension for open-weight models: the same model that can be steered by benign operators toward safer behavior can often be steered by malicious users toward disallowed behavior using similar interventions. In our setup, both directions are achievable via short prompt suffixes, as reflected by the large gaps $\Delta$ in Table~\ref{tab:suppression_results}. In real open-weight deployments, the attacker set is broader (e.g., fine-tuning, adversarial attacks and representation-level modifications), which can further narrow the practical separation between authorized and unauthorized steering. Improving this separation---reducing unauthorized steerability while preserving authorized steerability and utility---remains a central unsolved problem.

\section{Implications for Open-Weight Models}

The relationship between capability and control reveals a paradox. To avoid loss-of-control scenarios, we strive for high steerability with respect to \emph{safety-enforcing interventions}.  However, as of now, if an open-weight AI system is easily steerable by its builders, it is easily steerable by malicious actors \citep{qi2024_finetuning,arxiv2506_lowerssafety,arxiv2507_jailbreak_tuning}. If open-weight models approach "superintelligent" capabilities while maintaining high steerability, the dominant risk \citep{growiecprettner2025_pdoom} might not lie in ``uncontrollable AI'' but in large-scale human misuse. For open-weight systems, ``refusal training'' (teaching the model to decline harmful requests) is fragile  \citep{arxiv2502_geometry,openreview_mechanistic}. An attacker with white-box access can strip refusals via many techniques (e.g., fine-tuning, jailbreaking) \citep{arxiv2506_transferability,arxiv2506_suffixes}. Strictly preventing unauthorized steering in high-capability open-weight models remains an unsolved engineering problem.

Consequently, steerability is a double-edged sword. Safety benefits from high \emph{authorized steerability} (builders reliably reaching intended behaviors), while security benefits from low \emph{unauthorized steerability} (attackers failing to elicit disallowed behaviors). This creates a specific dilemma for open-weight models: releasing weights expands the attacker set (e.g., fine-tuning and weight editing), which increases unauthorized steerability. Existing proposals to address this dilemma target different levers, among others: \emph{capability removal} (unlearning/erasure), \emph{tamper resistance}, and \emph{interface restriction} (API-only deployment).

\paragraph{From refusal to unlearning.}
One idea is to pivot from behavioral refusals to capability unlearning (concept erasure). If a model's latent ability to represent a specific hazard (e.g., bio-weapon synthesis) is excised from the network weights, the system retains utility for general tasks while minimizing unauthorized steerability for that specific hazard. If successful, this makes the behavior significantly harder to elicit, as the model lacks the latent features required to represent the hazard \citep{arxiv2503_unlearning_survey,gandikota2024_elm,emnlp2025_erasure}.

\paragraph{Encrypted execution as mitigation.}
Another way to weaken the open-weight safety--security dilemma is to distribute a model in a form that is \emph{usable} but not practically \emph{modifiable}---i.e., users can run inference on their own infrastructure, but cannot trivially fine-tune or edit model. In principle, homomorphic encryption (HE) enables computation over encrypted values, which could protect model parameters; in practice, however, HE-based inference remains orders-of-magnitude slower than plaintext execution for nontrivial neural networks, and is widely viewed as impractical for large generative models as of today \citep{li2024_pti_survey,zhang2025_cipherprune,zhang2025_nexus}.

\paragraph{Fine-tuning-resistance (``fragile checkpoint'') immunization.}
Another immunization idea is to release a \emph{tamper-resistant} checkpoint: the model is intentionally placed in a parameter optimization region where small downstream fine-tuning attempts either (i) do not reliably achieve targeted behavioral changes, or (ii) trigger broad capability degradation (``capability collapse''). Intuitively, this aims to make attacker-driven gradient updates highly collateral by exploiting sharp curvature / poor conditioning so that moving toward a malicious objective breaks other competencies. Feasibility of these methods is for now moderate, so it is best treated as a deterrent layer rather than an ultimate defense. \citep{arxiv2408_tar,arxiv2412_durability,arxiv2502_tampering}

\paragraph{Current state.}
At present, the only widely deployed way to maintain high authorized steerability while reducing unauthorized steerability is to \emph{not} release model weights. Instead, users interact with the system through an API (optionally with server-side fine-tuning), where the provider retains control of the weights and the inference stack, can monitor for misuse, and can update mitigations over time. This does not eliminate \emph{black-box} unauthorized steering via prompting, but it materially reduces the attacker’s ability to remove safeguards by direct weight access (e.g., fine-tuning, representation engineering) which is usually easier and more effective than black-box adversarial techniques \citep{arxiv2412_durability,arxiv2508_worstcase,hoscilowicz2025adversarialconfusionattackdisrupting}.

\section{Limitations}

\texttt{InstrumentalEval} is small (76 scenarios), the judge evaluation is automatic, and the measured rates might be sensitive to non-conceptual prompt wording. Accordingly, we avoid over-interpreting small changes in results: differences of a few percentage points should be read as ``no clear shift under this protocol'' rather than strong evidence for or against the underlying tendency. The main signal we emphasize is the presence (or absence) of large, robust shifts (e.g., gaps on the order of tens of percentage points).

Moreover, presented experiments use the simplest steering technique to test whether instrumental-like behaviors are \emph{prompt-elicitable} and \emph{prompt-suppressible} \citep{arxiv2509_manipulating,universal_steering2025}. We expect that more sophisticated steering methods (e.g., additional post-training such as SFT/RL, or representation-level interventions such as activation steering / concept editing) \citep{arxiv2410_conceptors,arxiv2501_fgaa} could further increase or decrease measured convergence rates, and we leave systematic evaluation of such methods to future work.

\section{Conclusion}
We treat steerability as an empirical property: how strongly feasible interventions can shift model behavior under a stated elicitation and scoring setup. We emphasize that steerability is not a single axis. It splits into \emph{authorized steerability} (benign operators reliably enforcing intended behavior) and \emph{unauthorized steerability} (attackers eliciting disallowed behavior). Separating authorized from unauthorized steerability highlights a core tension for open-weight systems: the same mechanisms that enable reliable control by builders can also enable misuse by adversaries.

Using \texttt{InstrumentalEval}, we show that minimal inference-time steering (prompt suffixes) can strongly suppress or amplify outputs labeled as instrumental convergence under our protocol. In this regime, increased model capability does not imply weaker control; aligned variants remain highly responsive to constraint prompts, and scaling can coincide with lower convergence. Consequently, for open-weight models, preventing unauthorized steering remains a major open technical problem \citep{security_steerability2025,arxiv2412_durability,arxiv2310_managing}.

\bibliographystyle{unsrtnat}
\bibliography{_main}

\end{document}